\documentclass[letterpaper, 10 pt, conference]{ieeeconf}  

\IEEEoverridecommandlockouts                              

\usepackage[utf8]{inputenc}
\usepackage{leftidx}
\usepackage{bm}
\usepackage{amssymb,amsmath,amsfonts,empheq}
\usepackage{graphicx}
\usepackage{siunitx}
\usepackage{algorithm}
\usepackage[noend]{algpseudocode}
\usepackage{type1cm}
\usepackage{pifont}
\usepackage{lettrine}
\usepackage{color}
\usepackage{cite}
\usepackage{breqn}
\usepackage{tikz}
\usepackage{array}
\usepackage{import}

 \usepackage{microtype}

\usepackage{url}

\usepackage{multirow}

\global\long\def\norm#1{\left\Vert #1\right\Vert }%

\newcommand{\luis}[1]{{#1}}
\newcommand{\riddhiman}[1]{{\color{green!70!black!100}}}
\newcommand{\rafael}[1]{{\color{red!80!green!80}}}

\usepackage{graphicx}
\usepackage{amsmath}
\usepackage{float}
\usepackage[percent]{overpic}
\graphicspath{ {images/} }
\usepackage[bookmarks=true]{hyperref}

\IEEEoverridecommandlockouts                              

\title{\LARGE \bf
A Solution to Slosh-free Robot Trajectory Optimization }

\author{Rafael I. Cabral Muchacho, Riddhiman Laha, Luis F.C. Figueredo, and Sami Haddadin
\thanks{$^{1}$The authors are with Munich Institute of Robotics \& Machine Intelligence, Technische Universität München (TUM), Germany. 
  This work was funded by the Lighthouse Initiative Geriatronics by StMWi Bayern (Project X, grant 5140951), LongLeif GaPa gGmbH (Project Y, grant 5140953), ``Centre for Tactile Internet with Human-in-the-Loop'' (CeTI, grant 390696704)  
  and
  KI.FABRIK Bayern (grant DIK0249).
S. Haddadin has a potential conflict of interest as shareholder of Franka Emika GmbH. Email:\texttt{\{rafael.cabral, riddhiman.laha, luis.figueredo, haddadin \}@tum.de}
}
}

\begin{document}

\maketitle


\begin{abstract}
This paper is about fast slosh-free fluid transportation. Existing approaches are either computationally heavy or only suitable for specific robots and container shapes. We model the end effector as a point mass suspended by a spherical pendulum and study the requirements for slosh-free motion and the validity of the point mass model. In this approach, slosh-free trajectories are generated by controlling the pendulum's pivot and simulating the motion of the point mass. We cast the trajectory optimization problem as a quadratic program—this strategy can be used to obtain valid control inputs. Through simulations and experiments on a 7 DoF Franka Emika Panda robot we validate the effectiveness of the proposed approach.

%
%

\end{abstract}

\section{Introduction}

In this work, we are interested in the problem of 
transportation and optimal-manipulation of dynamic fluids and fragile materials. 
Particularly, we are focused on extreme time-optimal solutions that 
leverage robot capabilities to exceed human-level performance.  
Take for instance, the task of transporting a full cup of coffee, a glass of wine, 
or a hazardous liquid in industry. Although basic inspiration for designing such a system can be gained from innate human skills to carefully adjust the frequency of the containing cup/container to match that of the unrestrained free surface, achieving this in aggressive maneuvers for humans can be extremely hard.
Safety is the number one priority in this case, as no one likes to spill a coffee---neither do robots. Solutions often lie in minimizing higher derivative terms such as jerk and snap of the trajectories through optimal control strategies.
%
Taking inspiration from a market solution for coffee/liquid transportation as shown in Fig.~\ref{fig:big_picture}---namely the Spillnot mechanism \cite{spillnotPatent,spillnotTrademark}---we propose an optimization-based motion generation solution that provides 
above-human real-time fluid transfer at high speeds that predicts and compensates reaction forces through a simplified closed-form pendular-like dynamics---and thus complete the task without spilling liquids even for aggressive motion profiles. 

\begin{figure}[t]
    \centering
    \includegraphics[width=\columnwidth]{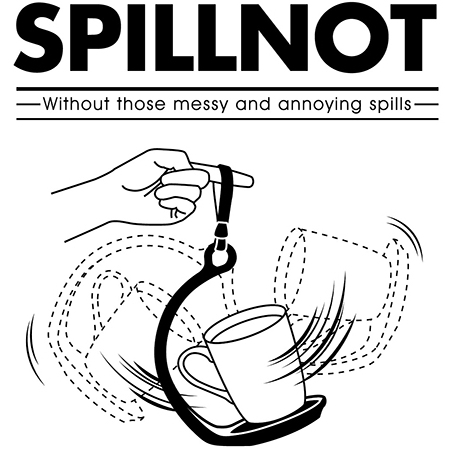} 
    \caption{Inspiration for our problem. Picture courtesy: Science Factory and SpillNot trademark \cite{spillnotPatent,spillnotTrademark}. See further details of the functioning of the the SpillNot mechanism \href{https://www.youtube.com/watch?v=qEDvmvBbDBk}{here}. 
    }
    \label{fig:big_picture}
\end{figure}

The problem of dynamic liquid behavior is well studied and dates back to Navier, Stokes~\cite{gilles2018navier}, 
the fluid packaging industry, and even the aerospace industry~\cite{Abramson66}.   
Motion induced sloshing is also a classic 
problem in control theory~\cite{grundelius2001methods, ibrahim2005liquid} wherein the main idea is to study the sloshing mode excited by the oscillations through resonance and additional frequency analysis \cite{zang2014slosh}. 
\luis{Among the robotics community, }while a plethora of approaches have been proposed 
to address the aforementioned transportation problem,  
\luis{most of them rely on motion optimization and/or smoothness to minimize jerks and acceleration leading to liquid sloshing, see for instance, \cite{liang2019making, reyhanoglu2012point}}.
\luis{A scarce amount of papers, have also addressed the problem through fluid analysis and optimization towards trajectory smoothing and container tilt-coupling, i.e., coupling with the trajectory \cite{aboel2009design, pridgen2013shaping}. Such approaches however often lead to computationally heavy, narrow and dedicated case-specific solutions.} 
\luis{Overall, existing frameworks 
addressing the the posed problem are either suboptimal, or computationally demanding in terms of solving the inverse motion problem.    
 Existing human-centred applications call for optimal, safe and mostly real-time solutions, that make use of existing collaborative robots capabilities and inner control-loops running, for instance, at 1 KHz. Neglecting or overseeing the robot's capabilities hinders real-world applications where it is desirable to have a faster and more effective solution.
 }
%
%
%

\luis{Contrary to existing methods, this work instead adopts a fundamentally different approach, and proposes solving the spilling-free fluid transportation problem by formulating it as a linearized point mass system. 
Casting the problem like this
enables us to 
integrate the tilt of the container with the maximum possible task-space acceleration and jerk given by the system constraints. It also allows designing optimal motion generation through real-time quadratic programming 
and therefore is able to run within the robot's fast inner control loop for real-time applications.
}
%
%
To the best of our knowledge, this is the first work to propose a slosh-free manipulation solution 
based on a closed-form linearized yet accurate dynamics that account for translation and tilt-coupled dynamics. Overall, the proposed method allows for  real-time capable motion generation for aggressive fluid maneuvers along a given path. A set of experiments exploring the Panda robot arm capabilities are devised to highlight the efficiency of the proposed method. 


\section{Related Work}
In the literature for fluid dynamics planning and control, researchers often estimate the interacting fluid states by deploying complex methods such as computational fluid dynamics (CFD)~\cite{muller2003particle, djavareshkian2006simulation}. However, CFD simulators for even simple objects like coffee in a cup are usually computationally demanding and require hours to converge---deeming them infeasible for most planning and control applications in robotics. 
Numerical approaches for slosh modeling using a simplified version of the Navier-Stokes equations are usually decoupled in the sense that they involve an intensive study of the properties of the system and an analysis of its controllability \cite{grundelius1999motion, lan2018iterative, barbu1999control, ge2020controllability}. 
\begin{figure}[t]
    \centering
    \includegraphics[width=0.75\columnwidth]{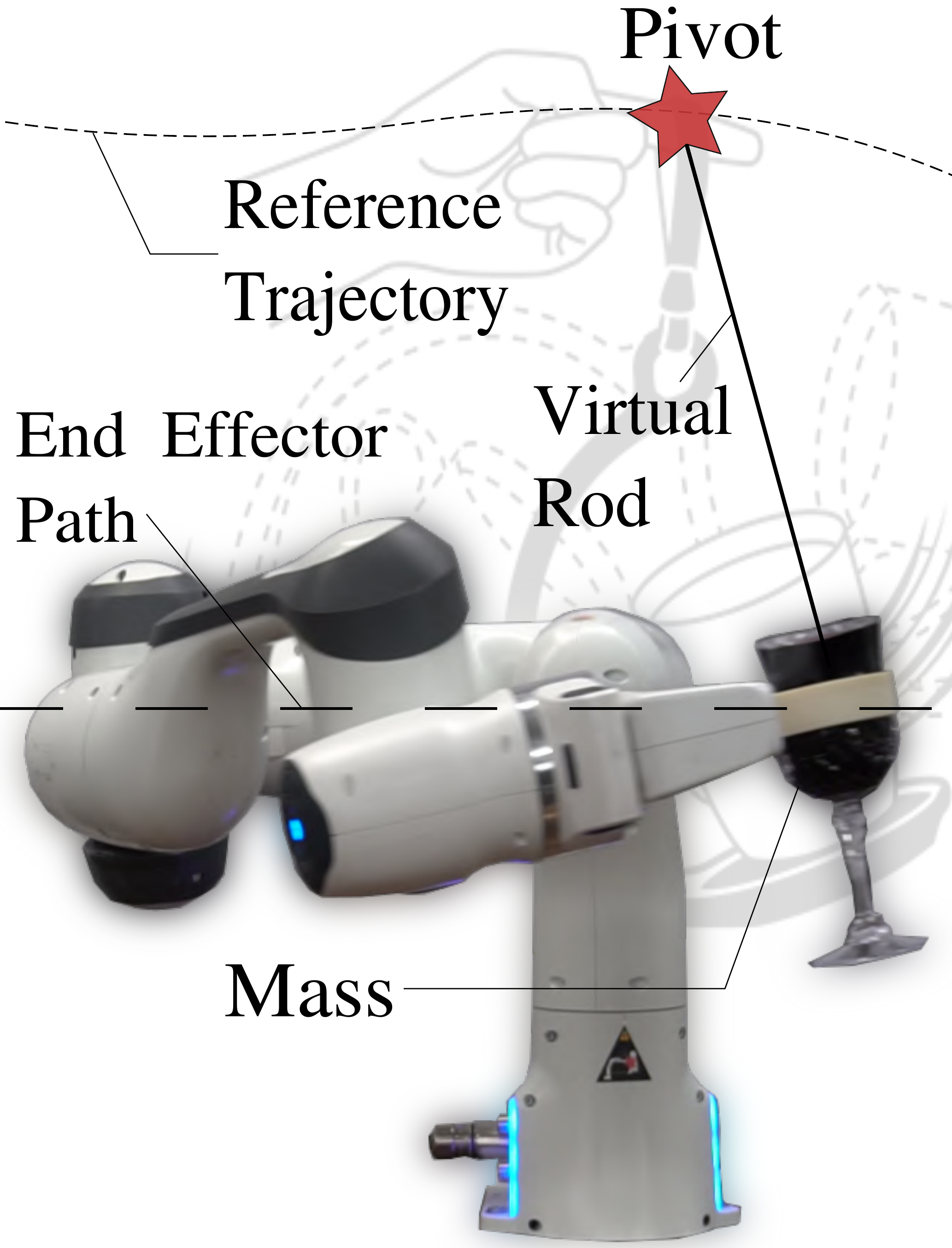} 
    \caption{Simplified dynamics model that we use for designing the optimization problem. The virtual pendulum forms the basis for analyzing the equations of motion. Constrained robot end-effector motion is achieved by controlling the pivot on the reference trajectory.
    }
    \label{fig:big_picture}
\end{figure}
The planar models \cite{wan2020waiter, yano2001liquid} are limited to modeling the s-velocity profiles for the underlying trajectory and designing time and frequency responses of the containing liquid; these solutions do not scale up to high DoF systems. 

\luis{As an alternative solution,}  researchers have also tried to model the system using approaches from Learning from Demonstration (LfD)~\cite{hartz2018pouring, rozo2013force}. The authors in \cite{pan2016motion} take an optimization approach to pouring, based on a simplified fluid dynamics model. 

In the realm of fluid manipulation, the two most important challenges are slosh-control~\cite{biagiotti2018plug, yano2001liquid} and pouring~\cite{schenck2018spnets, do2018accurate}. Therefore in this work, we exploit the capability of the robotic arms to change the orientation of the container similar to the works in \cite{feddema1997control, chen2007using, moriello2018manipulating}. The basic idea in these strategies is two fold: (a) apply smoothing methods~\cite{aboel2009design, pridgen2013shaping} to the desired trajectory, and (b) compensate for the expected remaining oscillations of the liquid by varying the container’s orientation. 
Furthermore, the novelty of our approach lies in the fact that we do not require any further knowledge about the robot except the provided reference trajectory and its kinematic limits. 
In this context, it is worth mentioning the works \cite{kuriyama2008trajectory,dang2004active}. 
In  \cite{kuriyama2008trajectory}, the proposed method explores the usage of evolutionary optimization methods to compute spill-free trajectories and impressively demonstrates the solution by transporting a filled spoon. The evaluation step of the algorithm is calculated through a detailed CFD simulation, which is accompanied by high computation expenses.
%
%
The work in \cite{dang2004active} relies  on direct and  
active compensation to 
investigate a 2D active control strategy to avoid spillage, sloshing and more generally to ensure a safe transport of delicate objects. To implement the control strategy they mount a parallel manipulator on the vehicle. The object is then carried by the manipulator and its motion “emulates a free swinging pendulum to minimize lateral forces acting on the object”. Regarding the length of the pendulum, instead of using a fixed value, they take it as a control variable. 
The modeling of pendulum-like dynamics for the
motion of liquids is also addressed at 
\cite{HAN201687_studyOnCoffeeSpilling}, which proposes human grasping strategies to better compensate oscillations and resonance of the modelled dynamics. A similar model can be used for the analysis of human motion through dynamic primitives; the authors of \cite{8793687} compare the human motion to rest-to-rest control strategies for the unconstrained manipulation of non-rigid objects.   
%

In this work, we also take inspiration from pendulum-like dynamics, yet differently than existing literature we focus on the linear approximation of the 3D pendulum to define the trajectory optimization problem as a quadratic-program, directly combining the smoothing and tilt coupling steps and allowing for the enforcement of additional constraints.

\section{Preliminaries and Concept Proposal}
This section presents the underlying intuition behind our approach, mathematical background and the fundamentals of the proposed motion generation and control problems. 

\begin{figure}[t]
    \centering 
        \begin{overpic}[width=0.75\columnwidth]{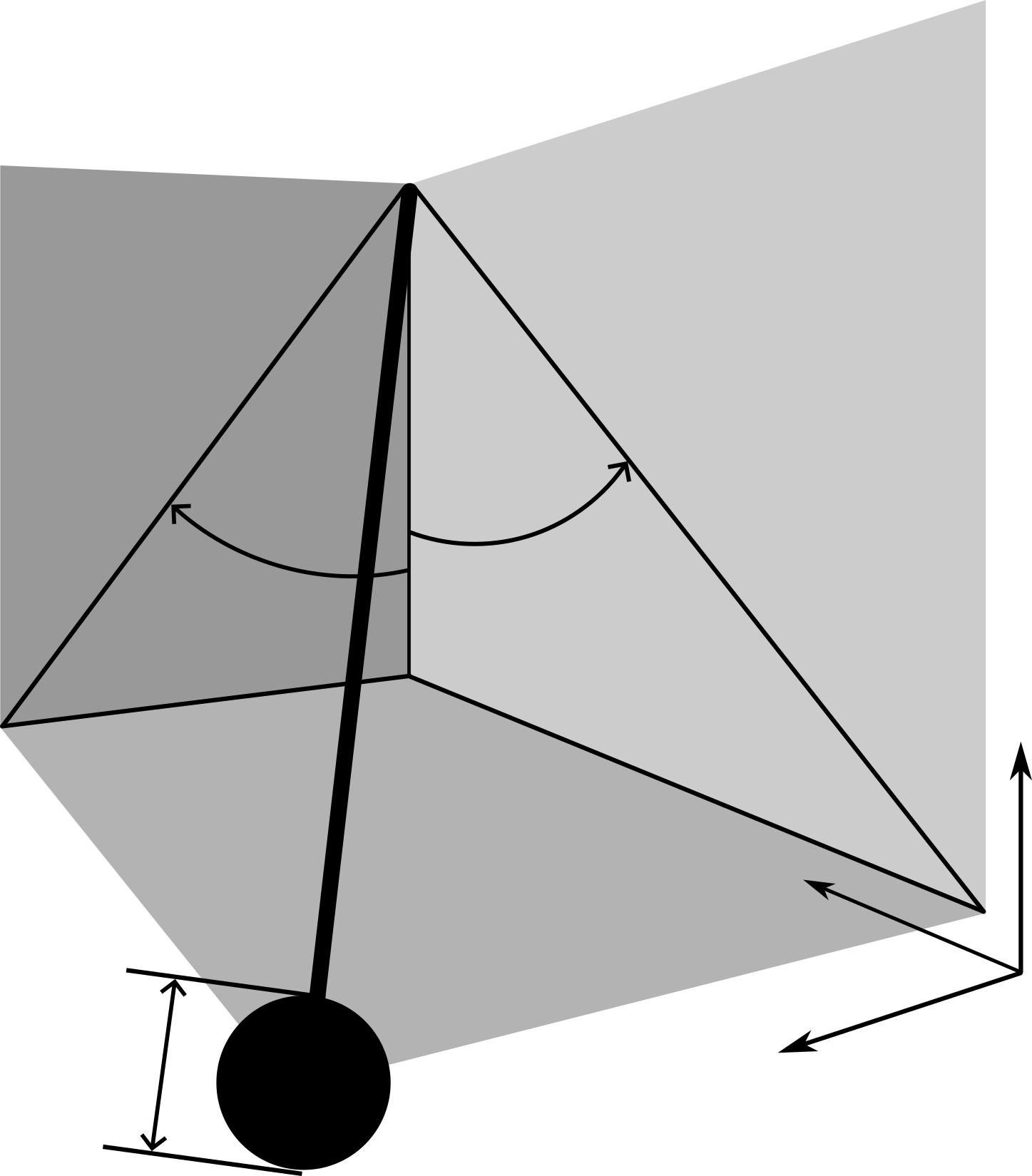}
    \put (22,55) {\huge$\displaystyle\phi$}
    \put (41,58) {\huge$\displaystyle\theta$}
    \put (7,7) {\huge$\displaystyle h$}
    \put (23,30) {\huge$\displaystyle l$}
    \put (60,8) {\huge$\displaystyle \hat{y}$}
    \put (62,22) {\huge$\displaystyle \hat{x}$}
    \put (81,35) {\huge$\displaystyle \hat{z}$}
    \put (35,3) {\huge$\displaystyle \bm{x}_m$}
    \put (28,89) {\huge$\displaystyle \bm{x}_p$}
    \end{overpic}

\begin{tabular}{l l | l l}
\\
\bf{Rod length} & $l$ & \bf{Object height} & $h$ \\
\bf{Pivot position} & $\bm{x}_p(t)$ & \bf{Mass position} & $\bm{x}_m(t)$ \\
\bf{Tilt angle ($\hat{y}$}) & $\theta(t)$ & \bf{Tilt angle ($\hat{x}$}) & $\phi(t)$ \\
\end{tabular} \\

    \caption{Spherical pendulum parametrization and nomenclature.}
    \label{fig:state_vars}
\end{figure}

\subsection{Conditions for Slosh-free Motion}\label{sec:sloshcond}
Intuitively, \luis{one can reason about the sloshing problem as being caused by the reaction forces exerted on the liquid from the inner walls of the container due to higher-derivatives impulse-like motions, 
for instance, 
high jerks and accelerations. 
Therefore,} a slosh-free motion can be found when there are no lateral reaction forces nor external torques acting on the container. This case arises when the total reaction force is aligned with object's vertical direction and its line of action passes through the object's center of mass. The relationship between the lateral acceleration $a_{x}$, vertical acceleration $a_{z}$, the gravity acceleration $\mathrm{g}$ and the tilt angle $\theta$ is therefore given by~\cite{moriello2018manipulating}
\begin{equation}
\label{eq:slosh_free_conditions}
(a_z + \mathrm{g})\tan\theta = a_x.
\end{equation}
The above relationship builds the base of our motion generation 
the presented conditions as performance measures.

\subsection{Spherical Pendulum Parametrization}
\label{subsec: spherical_pend_prmtzn}
The motion of a planar pendulum---based on a movable pivot---is trivial to obtain and can be fully described by 
the position of the pivot, the tilt angle and their derivatives. 
On the contrary, the $3$D pendulum case---also called spherical pendulum---has no similar universal parametrization. 
%


Parametrizations can be based on unit-dual quaternions---a rigid-body transformation representation that we also exploit in our robot modelling. Dual-quaternion algebra has certain advantages over other representations (e.g., translation and attitude coupling, singularity-free representation and computational efficiency \cite{Laha_Thesis,1990_Funda_Paul_TRA,OZGUR201666}) and frameworks\cite{1988_Khatib_ISRR,1988_Uchiyama_Dauchez_ICRA,1994_Connolly_Pfeiffer_CDC}. For further information about the Lie group of unit-dual quaternions applied to control we refer readers to \cite{2016_Figueredo_PhDThesis,FIGUEREDO2021109817,adorno2017robot}. This representation is singularity-free and can therefore be useful for tasks where all orientations or sphere segments are likely to be visited. 

The parametrization that we use in this work are based on the linear model of the pendulum (see Fig. \ref{fig:state_vars}). The capabilities of an optimization based approach are tested while being careful that the assumption of limiting the tilt angles for the linearization holds. As a consequence of the small angle approximation and linearization around the stable equilibrium, minimal parametrizations without singularities near the equilibrium point will be beneficial for the task. The widely used spherical coordinates are singular in the “poles” and are therefore not adequate. Another set of minimal coordinates describes the angles of two planar pendulums orthogonal to each other. This representation is not singular near the bottom equilibrium point. Applying the superposition property of linear systems it follows that the resulting equations can be described by overlapping the resulting motion of the two orthogonal simple pendulums \cite{campa2021modelling}. The parametrization and notation of choice can be seen in Fig. \ref{fig:state_vars}.

\subsection{Equations of Motion}

In this subsection, we use the Euler-Lagrange equations to derive the equations of motion of the spherical pendulum based on the chosen parametrization in~\ref{subsec: spherical_pend_prmtzn}. The rod of length $l$ is taken as weightless and no friction sources affect the motion.

As a function of the variables and parameters (\ref{fig:state_vars}), the position of the mass can be described by

\begin{equation}
    \bm{x}_m = [ \ x \ y \ z \ ]^{T} = \begin{bmatrix}
     x_p - l\sin{\theta} \\
           y_p + l\cos{\theta}\sin{\phi} \\
           z_p - l\cos{\theta}\cos{\phi} \\           
    \end{bmatrix}.
    \label{eq:system state r}
\end{equation}

The mass of the system is concentrated at this point and the total energy is therefore determined by the motion therein—the evolution of $\bm{x}_m$ with time. 
The kinetic energy $K$ and potential energy $U$ of the system can be calculated as
\begin{align}\label{eq:cooperative variables}
    K &= \frac{1}{2}m\dot{\bm{x}}_m^T\dot{\bm{x}}_m, \  \quad \text{and}  \  \quad
    U = m \mathrm{g} z.
\end{align}
Building the Lagrangian $L:=K-U$ and applying the Euler-Lagrange equation gives the equations of motion of the system,

\begin{align}\label{eom}
    l\ddot{\theta} =&  - \sin{\theta}(\mathrm{g} + \ddot{z}_p)\cos{\phi} + \ddot{x}_p\cos{\theta}  \nonumber \\ 
    & + \ddot{y}_p\sin{\phi}\sin{\theta} - l\cos{\theta}\sin{\theta}\dot{\phi}^2 \\ \nonumber \\
    l\cos{\theta}\ddot{\phi} =& - \sin{\phi}(\mathrm{g} + \ddot{z}_p) -\ddot{y}_p\cos{\phi} + 2l\dot{\phi}\dot{\theta}\sin{\theta} .
\end{align}
\\
%
Further investigation of the nonlinear equations 
show that the pendulum motion fulfills the slosh-free condition presented in~\ref{sec:sloshcond}.

\subsection{Validity of a Point Mass Model}\label{sec:pmm}

Due to the complexity of accurate modeling of fluid dynamics, we model the fluid as a point mass. In this section, we investigate the conditions which lead to a valid point mass approximation of an object. A point mass model of a rotating object is appropriate if the mass moment of inertia with respect to its center of mass is negligible relative to the mass moment of inertia with respect to the rotation axis. We introduce the approximation error $p$ which is non-negative and serves as a measure for the validity of the approximation. The following relationship 
\begin{align}
(1+p)  l^2  m &= l^2m + J_c, \\
p  l^2  m &= J_c ,
\end{align}
relates the approximated mass moment of inertia $ml^2$ with the actual mass moment of inertia---$J_c$ with respect to the center of mass and $ml^2$ from the parallel axis theorem---through the error factor $(1+p)$. We can see that as $l$ increases, $p$ approaches $0+$ asymptotically. We can model $J_c$ as the mass moment of inertia pertaining to a cube with sides $h$ equal to the length of the object ($J_c = mh^2/6$) which is a higher bound on $J_c$. Simplifying we obtain
\begin{align}
l = \frac{h}{\sqrt{6p}}, \quad \mathrm{or} \quad  r = \frac{1}{\sqrt{6p}}, 
\end{align}
 where $h$ is known from the object’s geometry. The validity of the approximation can be expressed as a function of the ratio $r$, which represents the $l$-to-$h$ ratio. As a rule of thumb, a ratio $r=l/h=3$ leads to an approximation error of $p<1.8\%$.


\section{SpillNot Motion-Generator Controller}



\luis{
The proposed motion generation control scheme relies on 
a quadratic programming (QP) formulation of 
the pendulum-like dynamics described in the previous section.
%
To model the nonlinear spherical pendulum dynamics and
devise a QP-based solution to the predictive motion,
herein we explore a linearized model of the system and discuss the consequences and limitations of the approach. 
}
%
First, we present the first order approximation by Taylor expansion of \eqref{eom} with respect to the state variables at the stable equilibrium,  
$ \theta=\phi=\dot{\theta}=\dot{\phi}=\dot{x}_p=\dot{y}_p=\dot{z}_p=0    $. 
This 
results in 
\begin{align}
    \dot{\bm{x}} &= [ \ \dot{x}_p \ \dot{y}_p \ \dot{z}_p \ \dot{\theta} \ \dot{\phi} \ u_1 \ u_2 \ u_3 \ \ddot{\theta} \ \ddot{\phi} \ ]^T
\end{align}
where 
\begin{align}
    \bm{u}        &= [ u_1 \ u_2 \ u_3]^T = [ \ \ddot{x}_p \ \ddot{y}_p \ \ddot{z}_p \ ]^T ,\\
    l\ddot{\theta} &= -g\theta + u_1, \\
    l\ddot{\phi}   &= -g\phi - u_2.
\end{align}
As previously described in \eqref{eq:system state r},  
the position and velocity of the mass are given by $\bm{y} = [ \ \bm{x}^T_m \ \dot{\bm{x}}^T_m \ ]^T$ which are determined by the state variables of the system. 
Following a similar procedure as above, we obtained the first order approximation of $\bm{y}$ as follows

%
\begin{align*}
    \bm{y} & {=}\left[  
\begin{array}{cccccc}
    \left(x{-}l\theta\right) \negthickspace &
    \left(y{+}l\phi\right)\negthickspace & 
    \left(z{-}l\right) \negthickspace & 
    \left(\dot{x}{-}l\dot{\theta}\right)\negthickspace & 
    \left(\dot{y}{+}l\dot{\phi}\right) \negthickspace & 
    \dot{z}
\end{array}
\right]^T.
\end{align*}

We make use of the derived first order approximations and restate the relationships in a continuous linear state space model
\begin{align}
    \dot{\bm{x}} &= \bm{A}_c\bm{x} + \bm{B}_c\bm{u}\\
    \bm{y} &= \bm{C}_c\bm{x} + \bm{D}_c\bm{u}.
\end{align}
We then discretize the model with a given time step using zero order hold (constant inputs between nodes) and employ the discrete state space matrices $\bm{A},\ \bm{B},\ \bm{C},\ \bm{D}$ to define the optimization objective and constraints.
\subsection{The Quadratic-Programming Formulation}

The goal of the optimization-based motion generation is to devise a feasible trajectory which deviates the least from a desired nominal one, whilst adhering to the pendulum motion constraints and to the initial and final conditions of the state.  
In this case,  we consider the total deviation as the sum of squared distances between the desired and optimized positions.
The optimization cost $J$ is therefore defined as  
\begin{align}
    J &= \frac{1}{2} \sum^{N}_{k=0}{{\norm{\bm{C}\bm{x}_k - \bm{y}_{d, k}}^2}} ,
    \label{eq:optimziation QP}
\end{align}
where $N$ is the number of nodes in the discretized trajectory.


The objective function \eqref{eq:optimziation QP}  
can be computed  at each node of the optimization independently. 
Considering a single node, it follows
\begin{align}
        J_k &= \frac{1}{2}{{\norm{\bm{C}\bm{x}_k - \bm{y}_{d, k}}^2}} \\
         J_k &= \frac{1}{2}{\bm{x}_k}^T \bm{C}^T \bm{C} \bm{x}_k - {\bm{y}_{d, k}}^T \bm{C} \bm{x}_k + \frac{1}{2}{\bm{y}_{d, k}}^T {\bm{y}_{d, k}}\\
         J_k &= \frac{1}{2}{\bm{x}_k}^T \bm{C}^T \bm{C} \bm{x}_k - {\bm{y}_{d, k}}^T \bm{C} \bm{x}_k. 
\end{align}
In the last step we choose to drop the last term from the cost and redefine $J_k$ accordingly, given that it is a constant and does not alter the minimizer. Notice that the evaluation of the cost at each node is a quadratic function on the pendulum's state at the corresponding node. The full cost can be therefore represented as a quadratic function on the $\bm{\chi}$ vector,
\begin{align}
        J = \frac{1}{2} {\bm{\chi}}^T \bm{H} \bm{\chi} - {\bm{g}}^T \bm{\chi}, 
\end{align}
with 
\begin{align}
        \bm{\chi} &= [{\bm{x}_0}^T, \dots, {\bm{x}_N}^T, {\bm{u}_0}^T, \dots, {\bm{u}_{N-1}}^T]^T,\\[1.2ex]
        \bm{H} &= \mathrm{diag}( \{ \ \bm{C}^T \bm{C} \ \}_{i=0}^{N},  \ \bm{0}_{3N \times 3N} )\\[1.2ex] 
        \bm{g} &= [ \ ( \bm{Y}_d \bm{C} )_{s}^T \ \bm{0}_{1\times3N}^T \ ]^T
\end{align}\\
where $\mathrm{diag}$ is a block diagonal matrix from its elements, and the $s$ subscript on a matrix refers to the column vector composed by the vertically stacking the columns of the matrix. The pendulum state $\bm{x}$ and input $\bm{u}$ at each node are the decision variables of the optimization and are vertically stacked in the $\bm{\chi}$ vector. 
The vector $\bm{y}$ describes the position and velocity of the mass, whilst 
the desired trajectory is represented in the matrix $\bm{Y}_d$ 
and consists of vertically stacked $\bm{y}_d^T$ row vectors. 
Notice that the index $d$ always stands for desired.  
%


Herein, we enforce the pendulum dynamics as equality constraints. This results in ten equality constraints—from the size of the pendulum's state $\bm{x}$—for each trajectory segment, i.e. $10N$ equality constraints. 
The definition of these constraints is simplified by making use of the discrete state space model matrices, as 
\begin{align}
    \bm{A}\bm{x}_{k} + \bm{B}\bm{u}_k - \bm{x}_{k+1} = \bm{0}_{10\times1}, \quad k \in 0, \dots, (N-1), 
\end{align}

Limits on the state and control variables can be formulated as element wise inequality constraints 
\begin{align}
\bm{\chi} \leq \bm{ub}_{\bm{\chi}} \quad \mathrm{and} \quad -\bm{\chi} \leq \bm{lb}_{\bm{\chi}}  
\end{align}
where the vectors $\bm{ub}_{\bm{\chi}}$ and $\bm{lb}_{\bm{\chi}}$ respectively contain the upper and lower bounds to be enforced.

The rate of change of the inputs corresponds to the jerk of the pivot's motion. This quantity determines to a large extent the aggressiveness of the resulting trajectory. To define the constraints for this quantity we employ finite differences
\begin{align}
    \frac{1}{T_s}(\bm{u}_{k+1}-\bm{u}_{k}) & \leq \bm{ub}_{\dot{\bm{u}}}, \quad k \in 0, \dots, (N-2)\\
    -\frac{1}{T_s}(\bm{u}_{k+1}-\bm{u}_{k}) & \leq \bm{lb}_{\dot{\bm{u}}}, \quad k \in 0, \dots, (N-2)
\end{align}
where the vectors $\bm{ub}_{\dot{\bm{u}}}$ and $\bm{lb}_{\dot{\bm{u}}}$ respectively contain the upper and lower bounds of the pivot jerk to be enforced.

We can define additional equality constraints for the first and last node of the trajectory, in cases where one would like the first and final states of the optimized trajectory to exactly match the desired state or if for example a zero velocity constraint is necessary for the execution of the trajectory. In our case, we define 
\begin{align}
    \bm{C}\bm{x}_k - \mathrm{diag}(\mathbb{I}_{3 \times 3}, \ \bm{0}_{3 \times 3})\bm{y}_{d, k} = \bm{0}_{6\times1}, \quad k \in \{0, N\}
\end{align}
equality constraints to enforce an exact match in the first and last positions of the mass and further to enforce zero velocity and acceleration in these two states.
Similar equality constraints can be defined for arbitrary states in the trajectory, not only in the initial and final nodes. This can be useful in point to point trajectories where the resulting optimized trajectory should exactly pass through the defined points\cite{kelly2017introduction}\cite{betts2010practical}.

Putting it all together, we obtain

\begin{alignat}{2}
& \bm{\chi}* = &\operatorname*{argmin}_{\bm{\chi}} &       \frac{1}{2} {\bm{\chi}}^T \bm{H} \bm{\chi} - {\bm{g}}^T \bm{\chi}\label{eq:optProb}\\
 &\text{subject to} &      & \bm{A}\bm{x}_{k} + \bm{B}\bm{u}_k - \bm{x}_{k+1} = \bm{0}_{10\times1},\\
 & & & \bm{\chi} - \bm{ub}_{\bm{\chi}} \leq 0,\\
 & & & -\bm{\chi} + \bm{lb}_{\bm{\chi}} \leq 0,\\
  & & & \frac{1}{T_s}(\bm{u}_{k+1}-\bm{u}_{k}) \leq \bm{ub}_{\dot{\bm{u}}},\\
 & & & -\frac{1}{T_s}(\bm{u}_{k+1}-\bm{u}_{k}) \leq \bm{lb}_{\dot{\bm{u}}},\\
 & & & \bm{C}\bm{x}_k - \mathrm{diag}(\mathbb{I}_{3 \times 3}, \ \bm{0}_{3 \times 3})\bm{y}_{d, k} = \bm{0}_{6\times1}.
\end{alignat}

\subsection{Joint Space Trajectory}

The optimal $\bm{\chi}$ returned by the optimization contains the states of the virtual pendulum describing the slosh-free work space trajectory. To evaluate whether the trajectory can be executed by a given robotic manipulator we are interested in obtaining a joint space trajectory which adheres to the manipulator's kinematic and dynamic constraints and also accurately tracks the slosh-free trajectory. Given that the optimized trajectory is smooth and continuous we make use of differential inverse kinematics to calculate the joint space trajectory.

We first reconstruct the poses of the mass from the pendulum states as dual-quaternions \cite{2016_Figueredo_PhDThesis} and then employ the pseudo-inverse of the manipulator's zero-Jacobian \cite{10.5555/561828} to obtain the joint velocities at each state. 
The description of the coordinate invariance dual-quaternion Jacobian is given in \cite{2013_Figueredo_Adorno_Ishihara_Borges_ICRA}.  
Further we integrate the velocities to obtain the joint configurations and apply finite differences to obtain the accelerations and jerks describing the joint motion.
Note that the initial joint configuration is necessary and can be provided by the user or otherwise calculated through inverse kinematics. 
We proceed to calculate the joint torques of the trajectory by simulating a dynamic model of the manipulator and evaluate the given dynamic limits constraints \cite{10.5555/561828}.

\section{Simulation and Experiments}


 In this section, we observe the results of the presented trajectory optimization framework and evaluate the spill-free constraints for each optimized trajectory. The simulations and experiments are based on a standard torque controlled Franka Emika Panda Robot.

As a desired trajectory, we evaluate a position step of $0.3\mathrm{m}$ in $x$ direction of the robot's base frame. 
The following results show the optimized trajectories and their evaluations. The container is approximated by a cube of side length $0.1$m and $10$g load at the end-effector. The parameter $r$ represents the ratio of pendulum length to container length, as described in section \ref{sec:pmm}. The rod length in the optimization is set accordingly to obtain $r$ parameters of ${3, 6}$ and $9$.

\subsection{Motion}

\begin{figure}[h]
\centering
\includegraphics[width=1\linewidth]{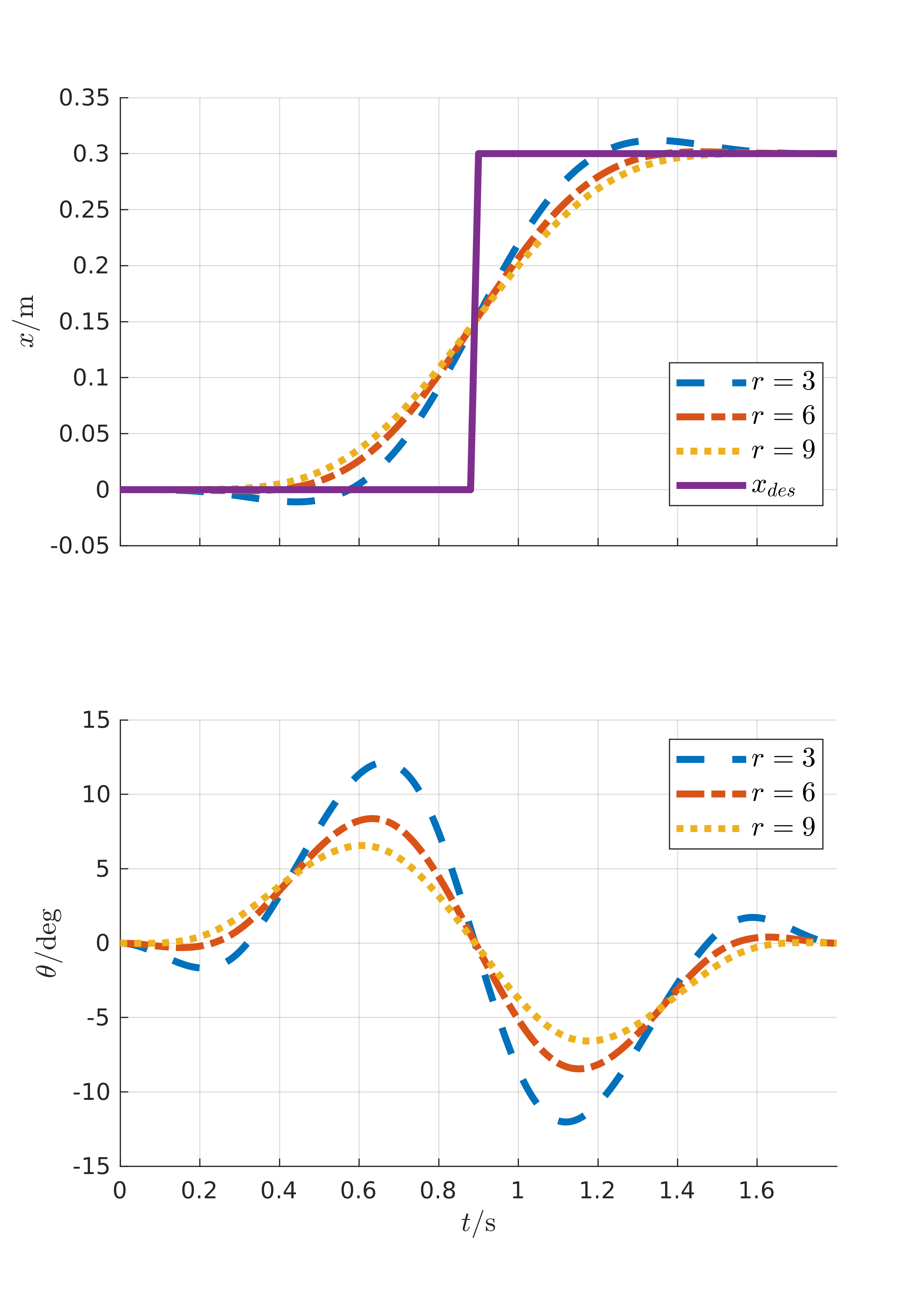}
\caption{Tilt angle (bottom) and translation (top) of the optimized trajectories with a desired step.}
\label{fig:pivot_traj}
\end{figure}

Figure \ref{fig:pivot_traj} shows the resulting translation and orientation of the optimized trajectories for different $r$ values. Lower $r$ can reach higher velocities at the cost of higher tilt angles and are therefore capable of obtaining more aggressive trajectories. The opposite can be said about higher $r$ values.

\subsection{External Forces}

\begin{figure}[t]
\centering
\includegraphics[width=1\linewidth]{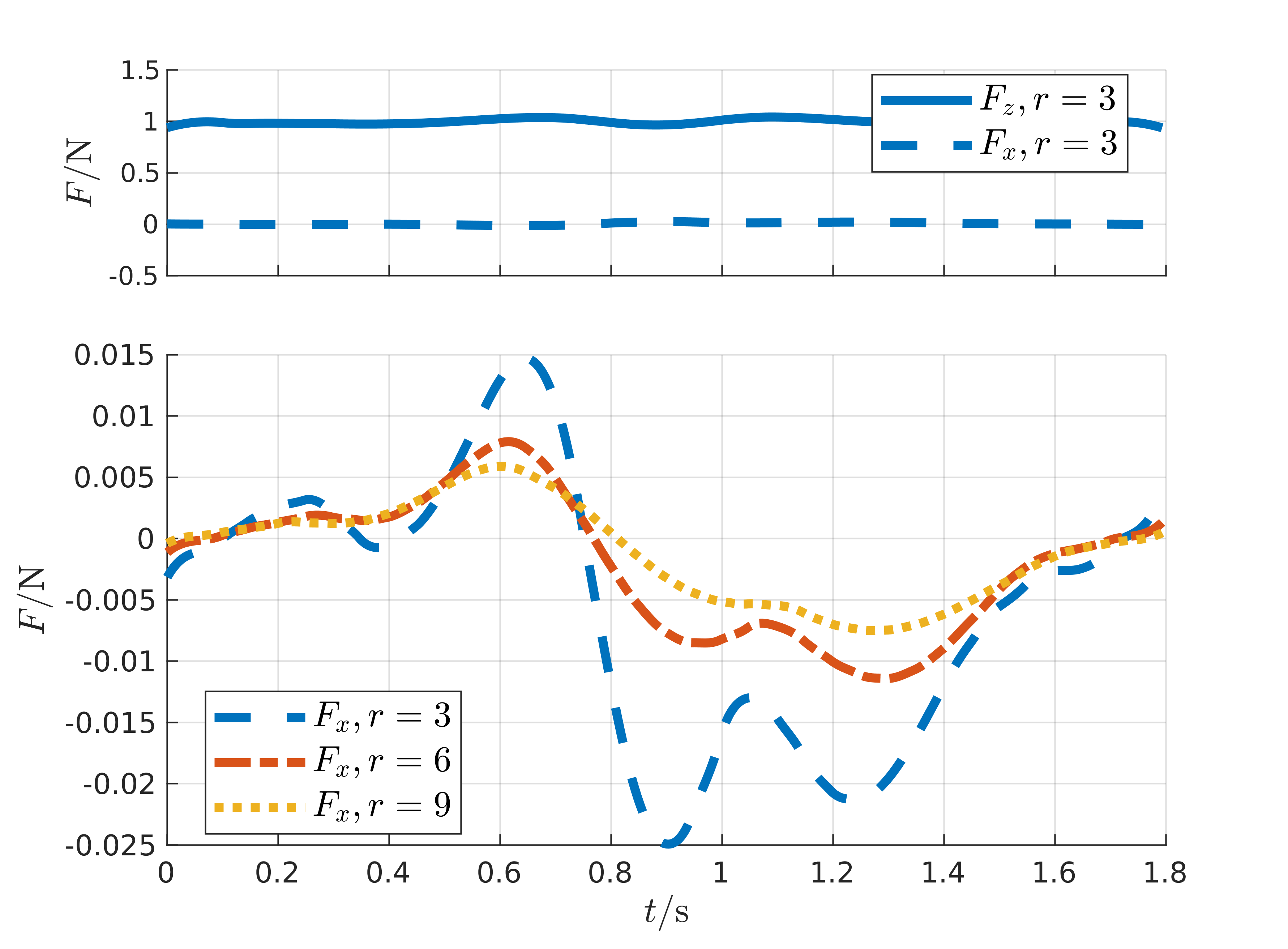}
\caption{$F_x$ lateral and $F_z$ vertical external forces exerted on the object by the manipulator during the optimized step motion.}
\label{fig:fx}
\end{figure}

After analyzing the resulting motion, we now address the dynamics of the trajectory, more specifically, the lateral external forces experienced by the object. Ideally—as described in section \ref{sec:sloshcond}—the forces acting on the object should be aligned with its vertical direction. In figure \ref{fig:fx} we show the magnitude of the simulated external forces along the axes of the end effector frame, where the object's vertical is aligned with the $z$-axis.

From figure \ref{fig:fx}, we can observe how the magnitude of lateral forces varies with a change in $r$. Higher $r$ values lead to lower magnitudes of lateral forces and the opposite can be stated as well. To which extent the kinematic and dynamic spill-free constraints are fulfilled is the topic of the following section.

\begin{figure*}[t]
\centering
\includegraphics[width=0.99\linewidth]{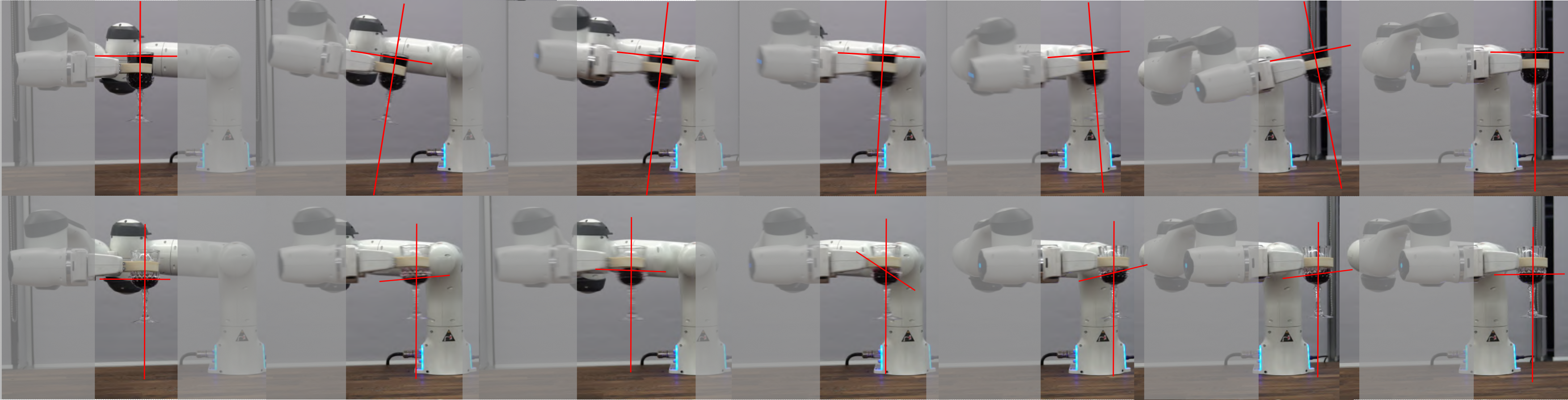} \vspace{-4pt} 
\caption{Top Panel: Slosh-free trajectory. Bottom Panel: Simple proportional controller for trajectory following results in spillage. As a reference, the maximum velocity of the presented motion is $0.63\mathrm{m/s}$. The red lines show the glass's vertical and the current orientation of the liquid's surface. The intersection of the lines at a right angle indicate slosh avoidance. 
}
\label{fig:Strip} \vspace{-17pt}
\end{figure*}
\subsection{Spill-free Constraint Evaluation}

Table \ref{table:csmeasures} shows the error of the kinematic and dynamic constraint measures of the tested trajectories. The force alignment error measures the maximum proportion of lateral forces to the total external force during the trajectory. The kinematic error refers to the maximum mismatch between the right and left side of the condition \eqref{eq:slosh_free_conditions} in Section \ref{sec:sloshcond}.

\begin{table}[b]
\centering
    \caption{Slosh-free constraint evaluation analysis be means of the worst-case ratio between lateral forces to the total external force during the trajectory and by means of the kinematic mismatch within the slosh condition in \eqref{eq:slosh_free_conditions}.}
    \label{table:csmeasures}
\begin{tabular}{ccc} 
 $r$ & Force Alignment Error & Kinematic Error  \\ 
 \hline
 3 & 2.51~e-02  & 7.66~e-04  \\ 
 6 & 1.13~e-02  & 4.54~e-04  \\ 
 9 & 0.75~e-02  & 3.15~e-04  \\ 
 \hline
    \end{tabular}
\end{table}
In all columns, the values decrease with an increase in $r$, due to the increase in accuracy of the point mass model and the decrease in maximum tilt, which in turn leads to a more accurate approximation by the linearized model. 




\subsection{Experiment}
To test our framework with real data we aimed to emulate an actual use case. We recorded a square-like path by demonstration and used this motion as a desired trajectory using a similar approach to \cite{laha2022coordinate}. Regarding the optimization parameters, the rod-length was set to $0.6\mathrm{m}$, the state boundary inequality constraints were set based on the Cartesian limits of the Panda Robot and we defined a time step of $33\mathrm{ms}$.
We executed the resulting joint space trajectory on the robot using a joint velocity controller, collected the trajectory data from the robot and evaluated the results. 

\begin{figure}[t]
\centering
\includegraphics[width=1\linewidth]{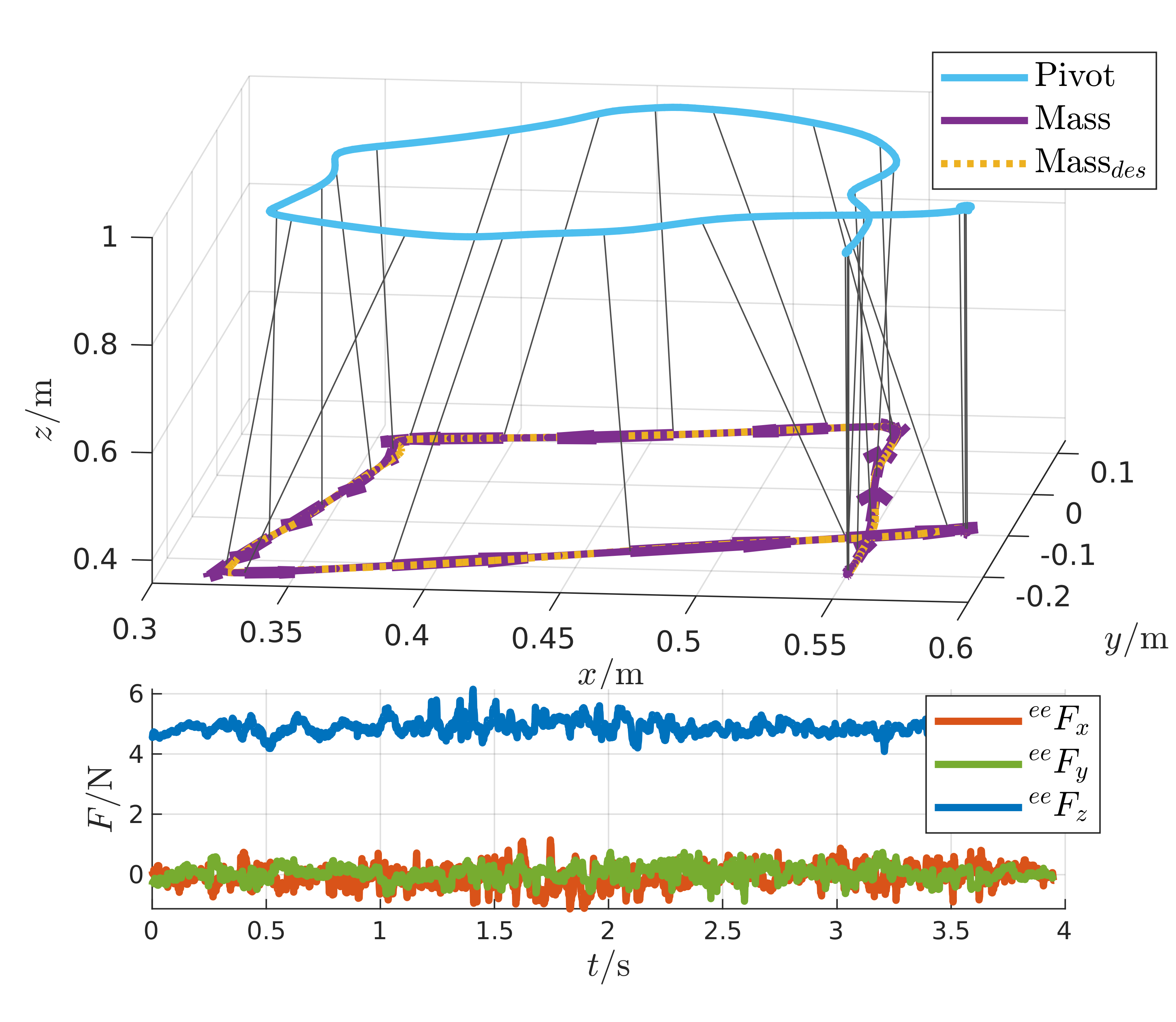}
\caption{Trajectory (top) and external forces (bottom) as a visualization of the 3D experiment on the Panda Robot. The dotted yellow line represents the trajectory demonstrated in the robot, which we used as a desired trajectory in the optimization framework. The external forces are measured from the robot torque sensors whilst executing the optimized trajectory.}
\label{fig:exp_square}
\end{figure}

Figure \ref{fig:exp_square} shows the resulting evolution of motion and external forces. From the top 3D-plot we can see that the optimized motion accurately matches the desired one. The thin black lines represent the rod of the virtual pendulum at different points throughout the trajectory and therefore show the tilt angle of the end effector. The bottom plot shows the external forces along the coordinates of the end effector frame, where the $z$-axis is aligned with the object's vertical direction. A maximum velocity of $0.48 \mathrm{m/s}$ and maximum acceleration of $1.22 \mathrm{m/s^2}$ were reached. The lateral forces remain close to zero during the trajectory, effectively avoiding spillage.

In a second experiment, the robot was commanded to follow a position step-like trajectory while grasping a glass filled with wine. For comparison, a simple proportional was used to move the end effector to the desired position. The results in form of a panel can be seen in figure \ref{fig:Strip}. The orientation of the wine's surface was constant relative to the glass throughout the optimized motion and spillage was successfully avoided.

\section{Conclusion}
In this work, we present a novel strategy for slosh-free liquid manipulation. Our approach exploits system linearization in addition to accounting for dynamic constraints of the robot. We notice that even for aggressive trajectories, the small angle approximation is rarely violated, which affirms the high value of the presented method for slosh-free trajectory generation.
Nevertheless, the limitations of the method include the fact that the point mass approximation depends on accurate knowledge of the center of mass. Trajectories with high vertical acceleration are not fully compensated, because the influence of $a_z$ is lost in the first order approximation. The object's yaw is currently held constant and limits the range of viable trajectories in joint space; expanding the spherical pendulum model to include yaw is trivial but the point mass approximation's validity becomes directly dependent on the mass moment of interia of the object around its vertical axis. 
We demonstrate that fluid simulation can be effectively avoided by investigating conditions which lead to approximately valid point mass models of the container. 
Using the state space form one can efficiently solve the trajectory optimization as a quadratic program.
In future work we plan to design a slosh-free cartesian PD-controller under interaction with environment, building on an admittance control framework. Additionally, because of the fast computation of a QP (limited nodes) the presented model can be used as part of a receding horizon control strategy for realtime reactive behavior.






\bibliographystyle{IEEEtran}
\bibliography{references.bib}

\end{document}